\documentclass[conference]{IEEEtran}
\usepackage{graphicx}
\usepackage{array}
\usepackage{adjustbox}
\usepackage[english]{babel}

\title{Automated Rating of Recorded Classroom Presentations using Speech Analysis in Kazakh}

\author{\IEEEauthorblockN{Akzharkyn Izbassarova, Aidana Irmanova and Alex Pappachen James}
\IEEEauthorblockA{
School of Engineering, Nazarbayev University, Astana\\
www.biomicrosystems.info/alex\\
Email:  apj@ieee.org}}

\begin{document}

\maketitle

\begin{abstract}

Effective presentation skills can help to succeed in business, career and academy. This paper presents the design of speech assessment during the oral presentation and the algorithm for speech evaluation based on criteria of optimal intonation. As the pace of the speech and its optimal intonation varies from language to language, developing an automatic identification of language during the presentation is required. Proposed algorithm was tested with presentations delivered in Kazakh language. For testing purposes the features of Kazakh phonemes were extracted using MFCC and PLP methods and created a Hidden Markov Model (HMM) \cite{c22, c22} of Kazakh phonemes. Kazakh vowel formants were defined and the correlation between the deviation rate in fundamental frequency and the liveliness of the speech to evaluate intonation of the presentation was analyzed. It was established that the threshold value between monotone and dynamic speech is 0.16 and the error for intonation evaluation is 19\%. 

\end{abstract}

\section{Introduction}

Delivering an effective presentation in today's information world is becoming a critical factor in the development of individuals career, business or academic success. The Internet is full of sources on how to improve presenting skills and give a successful presentation. These sources accentuate on important aspects of the presentation that grasps attention. 

Since there is no a particular template of an ideal oral presentation, opinions on how to prepare for oral presentations to make a good impression on the audience differ. For example, \cite{c2} claims that the passion about topic is a number one characteristic of the exceptional presenter. The author suggests that the passion can be expressed through the posture, gestures and movement, voice and removal of hesitation and verbal graffiti. Where the criteria for the content of presentation depend on the particular field, the standards for visual aspect and non-verbal communication are almost general for each presentation given in business, academia or politics. In the illustration of the examples of different postures and their interpretation the author emphasizes voice usage aspects like its volume, inflation, and tempo. It is important to mention that the author Timothy Koegel has twenty years of experience as a presentation consultant to famous business companies, politicians and business schools \cite{c2}. That is why the criteria for a successful presentation in terms of intonation given in this source can be used as a basis for speech evaluation as the the part of presentation assessment.

However, it can be questioned how the assessment of speech is normally conducted based on these criteria. \cite{c4} examined the different criterion-referenced assessment models used to evaluate oral presentations in secondary schools and at the university level. These criterion-referenced assessment rubrics are designed to provide instructions for students as well as to increase the objectivity during evaluation. It was suggested that intonation, volume, and pitch are usually evaluated based on the comments in criterion-referenced assessment rubrics like "Outstandingly appropriate use of voice" or "poor use of voice". The comments used in the evaluation sheets can be subjective \cite{c4} which is why the average relation between how people perceive the speech during the presentation and the level of change in intonation and tempo should be addressed.

In this paper we present a software for evaluating presentation skills of a speaker in terms of the intonation. We use the pitch to identify the intonation of the speech. Also, we aim to implement the automatic identification of the speech-language during the presentation as the presentations used for testing the proposed algorithm delivered in kazakh language. This task poses another problem, as Kazakh speech recognition is still not fully addressed in previously conducted research works. The recognition of the Kazakh speech itself is not within the scope of this paper. The adaptation of other languages such as Russian or English are considered as a next step.

The paper organized as follows: Section II presents the methodology of the design used for presentation evaluation, section III shows the results of testing the developed software and further section IV provides overall discussion of main issues of the software design.

\section{Methodology}

The Figure 1 illustrates the approach used to identify language and intonation. First, the features corresponding to the Kazakh phonemes are extracted.  Then the model for language recognition is developed based on Hidden Markov Model (HMM).

\begin{figure}
\centering
\includegraphics[width=0.48\textwidth]{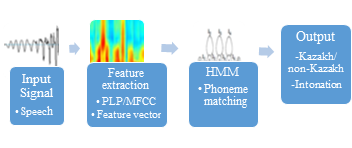}
\caption{Flow chart for speech evaluation}
\end{figure}

\begin{figure}
\includegraphics[width=0.5\textwidth]{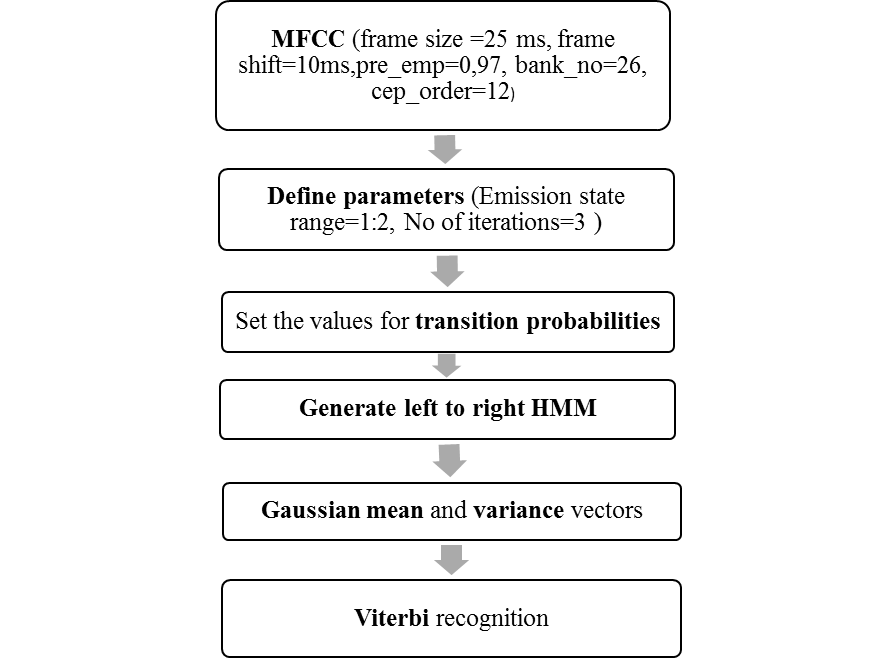}
\caption{Block diagram for phone recognition}
\label{blockdia}
\end{figure}

MATLAB is used to create a HMM for Kazakh phonemes. The block diagram in Fig. \ref{blockdia} illustrates the algorithm used in the code. 

The program should be able to evaluate the intonation and tempo of the speech. It is assumed that there is a direct correlation between the deviation rate in fundamental frequency and the liveliness of the speech. Thus, we need to conduct the pitch analysis to identify whether the proposed hypothesis is true. The pitch variation quotient derived from pitch contour of the audio files, where pitch variation quotient is a ratio of standard deviation of the pitch to its mean should be found. In order to identify the variation of pitch during presentations, the database of the presentations given in Kazakh language is created. This database consists of five presentations with ten-minute duration for each presentation. It is obtained by taking a video of students class presentations giving during "Kazakh Music History" and "History of Kazakhstan" courses at Nazarbayev University. For the simplicity of the analysis, presentations are divided into one-minute long audio files converted to WAV format. As a result, we obtain 32 audio files where seven presentations are with male voices and the rest by female. By using WaveSurfer program, the pitch value is found for each 7.5 ms of the speech. Two different sampling frequency values are tested to identify which sampling rate should be applied to obtain better results. 16 kHz and 44.1 kHz sampling frequency values are available in WaveSurfer. Thus, pitch is measured at these sampling rates. Then the mean and standard deviation of the pitch corresponding to each audio file is obtained. After that, a pitch variation quotient calculated. In order to obtain the pitch variation quotient we divide the standard deviation of the pitch to its mean. Finally, the results of the pitch variation quotient should be compared to the results of a perception test. 
The same speech files used for pitch extraction are used to conduct a test on how people perceive the speech regarding intonation. The purpose of this test is to identify the correlation between how people evaluate the presentation and the value of the pitch variation quotient. Since the paper aims to evaluate the presentation skills based on criteria such as intonation and tempo of the speech and give feedback to the users, the ability of the program to assess should be consistent with that how would professionals and general audience evaluate the presentation. Thus, we will ask students and professors to participate in this test. They will listen to a speech from presentations and categorize the speech into "monotone" or "emotionless" and "dynamic" or "lively". Since the intonation during the presentation is not always constant, the speech will be divided into small segments so the participants will give feedback for each speech segment. They should give marks for each presentations based on the intonation of the speakers. A marking system is a following: 1- monotone, 2- middle and 3-dynamic. After that, all results will be analyzed and the average mark for each presentation will be calculated. These average marks are compared with the results of the pitch variation quotient.

\section{Results} 
\subsection{Formants}

From data analysis results we defined first, second and third formants of Kazakh vowels. The Table 1 and Table 2 show the results for vowels produced by male and female voices, respectively. These phonemes were obtained by manually extracting each phoneme from KLC audio files.

\begin{table}[h]
\caption{Average formant frequencies of Kazakh vowels produced by male speakers}
\label{Average formant frequencies of Kazakh vowels produced by male speakers}
\begin{center}
\begin{tabular}{|c|c|c|c|}
\hline
Vowel & $F_1$, Hz & $F_2$, Hz & $F_3$, Hz\\
\hline

   \begin{minipage}{.05\textwidth}
   \centering
      \includegraphics[width=2.5mm, height=2.5mm]{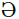}
    \end{minipage}& 734 & 1627& 2769\\ \hline
\begin{minipage}{.05\textwidth}
   \centering
      \includegraphics[width=2.5mm, height=2.5mm]{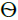}
    \end{minipage}& 517&1437&2500 \\ \hline
 \begin{minipage}{.05\textwidth}
   \centering
      \includegraphics[width=1mm, height=2.5mm]{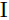}
    \end{minipage}& 540 & 1700&2705\\ \hline
 \begin{minipage}{.05\textwidth}
   \centering
      \includegraphics[width=2.5mm, height=2.5mm]{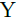}
    \end{minipage}& 513&1405&2505\\ \hline 
\begin{minipage}{.05\textwidth}
   \centering
      \includegraphics[width=2.5mm, height=2.5mm]{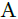}
    \end{minipage}& 811 &1258&2640\\ \hline
 \begin{minipage}{.05\textwidth}
   \centering
      \includegraphics[width=2.5mm, height=2.5mm]{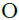}
    \end{minipage}&577&808&2765\\ \hline
 \begin{minipage}{.05\textwidth}
   \centering
      \includegraphics[width=2.5mm, height=2.5mm]{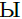}
    \end{minipage}& 590&1307&2652\\ \hline
 \begin{minipage}{.05\textwidth}
   \centering
      \includegraphics[width=2.5mm, height=2.5mm]{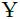}
    \end{minipage}&566&961&2605\\ \hline
\begin{minipage}{.05\textwidth}
   \centering
      \includegraphics[width=2.5mm, height=2.5mm]{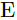}
    \end{minipage}&443&2087&2900\\ \hline
 
\hline
\end{tabular}
\end{center}
\end{table}

\begin{table}[h]
\caption{Average formant frequencies of Kazakh vowels produced by female speakers}
\label{Average formant frequencies of Kazakh vowels produced by female speakers}
\begin{center}
\begin{tabular}{|c|c|c|c|}
\hline
Vowel & $F_1$, Hz & $F_2$, Hz & $F_3$, Hz\\
\hline
 \begin{minipage}{.05\textwidth}
   \centering
      \includegraphics[width=2.5mm, height=2.5mm]{ae.png}
    \end{minipage}& 858 & 1929& 3180\\ \hline
 \begin{minipage}{.05\textwidth}
   \centering
      \includegraphics[width=2.5mm, height=2.5mm]{uo.png}
    \end{minipage}& 662&1424&2892 \\ \hline
 \begin{minipage}{.05\textwidth}
   \centering
      \includegraphics[width=1mm, height=2.5mm]{i.png}
    \end{minipage}& 697 & 1844&2986\\ \hline
 \begin{minipage}{.05\textwidth}
   \centering
      \includegraphics[width=2.5mm, height=2.5mm]{y.png}
    \end{minipage}& 572&1529&2801\\ \hline 
 \begin{minipage}{.05\textwidth}
   \centering
      \includegraphics[width=2.5mm, height=2.5mm]{a.png}
    \end{minipage}& 948 &1397&3048\\ \hline
 \begin{minipage}{.05\textwidth}
   \centering
      \includegraphics[width=2.5mm, height=2.5mm]{o.png}
    \end{minipage}&583&969&3220\\ \hline
 \begin{minipage}{.05\textwidth}
   \centering
      \includegraphics[width=2.5mm, height=2.5mm]{yi.png}
    \end{minipage}& 743&1175&3072\\ \hline
 \begin{minipage}{.05\textwidth}
   \centering
      \includegraphics[width=2.5mm, height=2.5mm]{uy.png}
    \end{minipage}&696&1116&3155\\ \hline
 \begin{minipage}{.05\textwidth}
   \centering
      \includegraphics[width=2.5mm, height=2.5mm]{e.png}
    \end{minipage}&554&2559&3150\\ \hline
 
\hline
\end{tabular}
\end{center}
\end{table}

The data given in Table 1 and Table 2 are used to observe the position of vowels according to their first and second formants. Figure 3 and Figure 4 illustrate the distribution of vowels for male and female voices respectively. 

\begin{figure}
\includegraphics[width=0.5\textwidth]{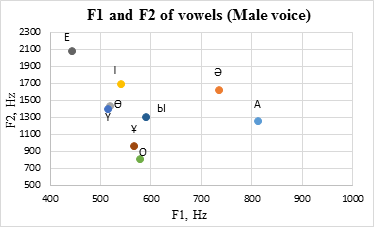}
\caption{First and second formant frequencies of Kazakh vowels produced by male speakers}
\end{figure}
\begin{figure}
\includegraphics[width=0.5\textwidth]{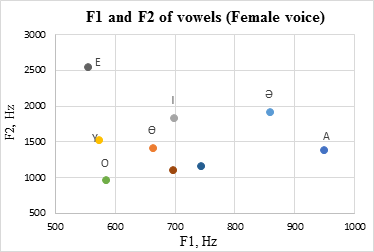}
\caption{First and second formant frequencies of Kazakh vowels produced by female speakers}
\end{figure}

\subsection{Intonation evaluation}
The test was conducted in order to identify how listeners perceive presentations based on intonation. Totally, 32 fragments from the different presentations given in the Kazakh language were tested. The participants of the test were ranking presentations from 1 to 3, where 1 is for monotone presentation and 3 is for dynamic. In addition, the variation of pitch in each presentation was measured and the pitch variation quotient was found. The pitch was measured for the different values of the sampling frequency. The average value for pitch variation quotient at f=16 kHz is 0.32 and at f=44.1 kHz the average quotient for 32 presentation fragments is 0.16. Figure 5 and Figure 6 show the results for pitch variation quotient of each presentation and their corresponding average marks based on the test results. Since the presentation were marked from 1 to 3, the average mark is 2. Thus, the boundary between monotone and dynamic presentation should be 2 along the x-axis and the average pitch variation quotient along the y-axis. In order to estimate error, the number of presentations with the value of pitch variation quotient below the average but with high average marks and inversely, the numbers of presentations with high pitch variation but low marks should be calculated. It is found that at f=16 kHz sampling frequency the error is 34\% and at f=44.1 kHz estimated error is 19\%.

\begin{figure}
\includegraphics[width=0.5\textwidth]{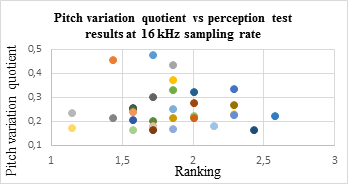}
\caption{Pitch variation quotient vs perception test results at 16 kHz sampling rate}
\end{figure}

\begin{figure}
\includegraphics[width=0.5\textwidth]{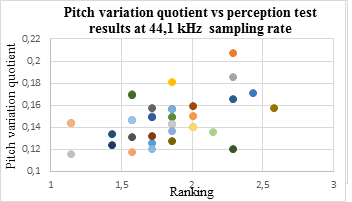}
\caption{Pitch variation quotient vs perception test results at 44.1 kHz sampling rate}
\end{figure}

Finally, the same presentation was recorded twice but with different intonations of the speech. The pitch variation quotient of the monotone speech is 0.092 whereas the second record with more dynamic intonation has 0.179 pitch variation quotient.

\subsection{Phone recognition}
As phone recognition does not recognize the speech, there is no need to use the lexical decoding, syntactic and semantic analysis. Therefore, phonemes are used as matching units. In this paper training the Kazakh phonemes for further phone recognition\cite{c29} was conducted in  MATLAB. The results are given from simulations of HMM with 1-emission and with 2-emission states. Models of context-independent phones which are represented by one or two emission states are shown in Figures 7 and 8, where $a_{ij}$ is a transition probability from state $i$ to $j$, while $S1...S4$ are transition states, $b_i(O_i)$ is probability density function for each state or emission probability, $O_i$ are observations.In Figure 7 $S1$ is an initial state, $S3$ is an end state and $S2$ is an emission state (Figure 7). For 2-emission state HMM, $S2$ and $S3$ represent emission states (Figure 8).

\begin{figure}
\includegraphics[width=0.48\textwidth]{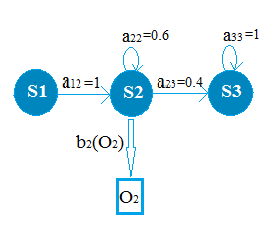}
\caption{1-emission state HMM}
\end{figure}

\begin{figure}
\includegraphics[width=0.5\textwidth]{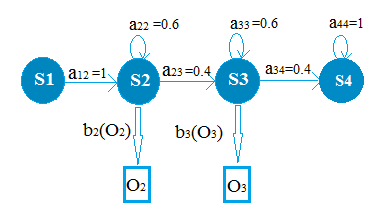}
\caption{2-emission state HMM}
\end{figure}

The phonemes recognition rate is calculated using Viterbi algorithm. Different sets of simulations are done with the variation of train and test data. Table 3 gives the results for recognition rates for 1-emission and 2-emission state. Train and test data contain phonemes recorded by female and male voices.

\begin{table}[h]

\caption{Recognition rate for 1-emission and 2-emission state HMM}
\label{Recognition rate for 1-emission and 2-emission state HMM}
\begin{center}
\begin{adjustbox}{center, width=0.45\textwidth}
\begin{tabular}{|c|c|c|}
\hline
Train/Test & Recognition rate for& Recognition rate for\\
 &1-emission state HMM &2-emission state HMM \\ \hline
Female/Female&61.76&64.71\\ \hline
 Male/Male&5.88&8.82\\ \hline
 Male/Female&11.76&14.71\\ \hline
 Female/Male&5.88&8.82\\ \hline 

\end{tabular}
\end{adjustbox}
\end{center}
\end{table}

\section{Discussion}

MFCC and PLP coefficients were extracted to develop phoneme based automatic language identification\cite{c18}. As a result, 12 cepstral coefficients and one energy feature were obtained for each feature extraction technique \cite{c18, c27}. After that, the first and second derivatives of these 13 features were taken , which gives 39- dimensional feature vector per frame in total to represent each phoneme.After that mean and covariance vectors for each phoneme were calculated. These values were used to create training model for the Kazakh phonemes recognition. MATLAB code was used to train the phonemes and create an HMM for them. As results show, the 2-emission state HMM gives higher recognition rate comparing with 1-emission state. In order to train for Kazakh language identification, the Kazakh corpus with labeling on phoneme level should be used. However, nowadays the word-level labeling is available in the current Kazakh Language Corpus\cite{c7}. This limits further analysis for phone recognition and language identification. More time is required to create a corpus with phoneme labeling. In this paper, we analyzed the Kazakh phonemes by extracting them manually in Praat program from the set of recordings done in a soundproof studio as well as in real environment conditions. For the Kazakh language identification based on the phonological features of the language itself, a bigger phoneme database is required.

\section{Conclusion}

To conclude, in this paper we present the system that can be used to evaluate presentation skills of the speaker based on the intonation of the voice. To test the proposed design we used data in kazakh language which consequently led to consideration of language identification system. As language identification and speech recognition is a relatively new field for Kazakh language processing field, we believe that the development of such system could be useful for the further popularization of Kazakh language and realization of different projects that builds up on top of the Kazakh speech recognition systems.

Future works cover the development of the Kazakh language corpus with the analysis and labeling up to phoneme level. After that, the language model for the Kazakh language can be developed. Finally, the larger database of the presentations in the Kazakh language should be created to analyze the presentation styles in the Kazakh language as well as to conduct a test and design an intonation evaluator.

\addtolength{\textheight}{-12cm}

\end{document}